\documentclass{article} 
\usepackage{nips14submit_e,times}
\usepackage{hyperref}
\usepackage{url}
\usepackage{amssymb,amsmath,amsthm}
\usepackage{subfigure}
\usepackage{tikz}
\usepackage{graphicx}
\usepackage{enumitem}
\usepackage{multirow}
\usepackage{algorithm,algorithmic}
\usetikzlibrary{positioning,decorations.pathreplacing}

\def\PP{{\mathcal P}}

\def\NN{{\mathcal P}'}
\def\HH{{\mathcal H}}
\def\Real{{\mathbb R}}
\def\x{{\mathbf x}}
\def\W{{\mathbf W}}
\def\y{{\mathbf y}}
\def\w{{\mathbf w}}

\def\u{{\mathbf u}}
\def\z{{\mathbf z}}
\def\tildephi{{\tilde{\varphi}}}
\def\tildepsi{{\tilde{\psi}}}
\def\defin{:=}
\newtheorem{definition}{Definition}
\newtheorem{lemma}{Lemma}
\newcommand{\normH}[1]{\left\|{#1}\right\|_{\HH}}
\newcommand{\norm}[1]{\left\|{#1}\right\|}

\newcommand{\normHz}[1]{\left\|{#1}\right\|_{\HH_0}}
\newcommand{\normE}[1]{\left\|{#1}\right\|_2}
\def\eg{{\it e.g.}}

\def\etab{\boldsymbol\eta}
\def\vs{\vspace*{-0.25cm}}
\def\vsb{\vspace*{-0.1cm}}
\def\kmone{k\text{--}1}

\title{Convolutional Kernel Networks}

\author{
   Julien Mairal, Piotr Koniusz, Zaid Harchaoui, and Cordelia Schmid \\
   Inria\thanks{LEAR team, Inria Grenoble, Laboratoire Jean Kuntzmann, CNRS, Univ. Grenoble Alpes, France.} \\
   \texttt{firstname.lastname@inria.fr}
}

\nipsfinalcopy 

\begin{document}

\maketitle

\begin{abstract}
   An important goal in visual recognition is to devise image representations that
are invariant to particular transformations. In this paper, we address this
goal with a new type of convolutional neural network (CNN) whose invariance is
encoded by a reproducing kernel. Unlike traditional approaches where neural
networks are learned either to represent data or for solving a classification
task, our network learns to approximate the kernel feature map on training data.

Such an approach enjoys several benefits over classical ones.  First, by
teaching CNNs to be invariant, we obtain simple network architectures that
achieve a similar accuracy to more complex ones, while being easy to train and
robust to overfitting. Second, we bridge a gap between the neural network
literature and kernels, which are natural tools to model invariance.  We
evaluate our methodology on visual recognition tasks where CNNs have proven to
perform well, \eg, digit recognition with the MNIST dataset, and the more
challenging CIFAR-10 and STL-10 datasets, where our accuracy is competitive
with the state of the art.

\end{abstract}

\section{Introduction}
We have recently seen a revival of attention given to convolutional neural
networks (CNNs)~\cite{lecun1998} due to their high performance for large-scale
visual recognition tasks~\cite{donahue2013,krizhevsky2012,wan2013}. The
architecture of CNNs is relatively simple and consists of successive layers
organized in a hierarchical fashion; each layer involves convolutions with
learned filters followed by a pointwise non-linearity and a downsampling
operation called ``feature pooling''. The resulting image representation 
has been empirically observed to be invariant to image perturbations and to
encode complex visual patterns~\cite{zeiler}, which are useful properties for
visual recognition. Training CNNs remains however difficult since high-capacity
networks may involve billions of parameters to learn, which requires both high
computational power, \eg, GPUs, and appropriate regularization
techniques~\cite{goodfellow2013,krizhevsky2012,wan2013}.

The exact nature of invariance that CNNs exhibit is also not precisely understood.
Only recently, the invariance of related architectures has been
characterized; this is the case for the wavelet scattering
transform~\cite{bruna2013} or the hierarchical models of~\cite{bou2009}.  Our
work revisits convolutional neural networks, but we adopt a significantly
different approach than the traditional one. Indeed, we use 
kernels~\cite{shawe2004}, which are natural tools to model invariance~\cite{decoste2002}.
Inspired by the hierarchical kernel descriptors of~\cite{bo2011}, we propose a
reproducing kernel that produces multi-layer~image~representations. 

Our main contribution is an approximation scheme called \emph{convolutional
kernel network} (CKN) to make the kernel approach computationally feasible. Our
approach is a new type of unsupervised convolutional neural network that is trained to approximate the
kernel map. Interestingly, our network uses non-linear functions that resemble
rectified linear units~\cite{bengio2009,wan2013}, even though they were not handcrafted
and naturally emerge from an approximation scheme of the Gaussian kernel map.

By bridging a gap between kernel methods and neural networks, we believe that
we are opening a fruitful research direction for the future. Our network is
learned without supervision since the label information is only used
subsequently in a support vector machine (SVM).  Yet, we achieve competitive
results on several datasets such as MNIST~\cite{lecun1998},
CIFAR-10~\cite{krizhevsky2009} and STL-10~\cite{coates2011} with simple
architectures, few parameters to learn, and no data augmentation.  Open-source
code for learning our convolutional kernel networks is available on the first
author's webpage.

\subsection{Related Work}\label{sec:related}
There have been several attempts to build kernel-based methods that
mimic deep neural networks; we only review here the ones that are most related to our
approach. 

\vs
\paragraph{Arc-cosine kernels.}
Kernels for building deep large-margin classifiers have been
introduced in~\cite{cho2010}. The multilayer arc-cosine kernel 
is built by successive kernel compositions, and each layer
relies on an integral representation.
Similarly, our kernels rely on an integral representation, and
enjoy a multilayer construction. However, in contrast to arc-cosine kernels:
(i) we build our sequence of kernels by \emph{convolutions}, using local
information over spatial neighborhoods (as opposed to compositions, using
global information); (ii) we propose a new training procedure for learning
a compact representation of the kernel in a \emph{data-dependent} manner. 

\vs
\paragraph{Multilayer derived kernels.}
Kernels with invariance properties for visual
recognition have been proposed in~\cite{bou2009}. Such kernels are built with a 
parameterized ``neural response'' function, which consists in computing the
maximal response of a base kernel over a local neighborhood. 
Multiple layers are then built by iteratively renormalizing the response kernels
and pooling using neural response functions.  Learning is performed by plugging
the obtained kernel in an SVM. In contrast to~\cite{bou2009}, we propagate
information up, from lower to upper layers, by using sequences of convolutions.
Furthermore, we propose a simple and effective data-dependent way to learn a
compact representation of our kernels and show that we obtain near state-of-the-art
performance on several benchmarks. 

\paragraph{Hierarchical kernel descriptors.}
The kernels proposed in~\cite{bo2011,bo2010} produce multilayer image
representations for visual recognition tasks.  We discuss in details these kernels in the next
section: our paper generalizes them and establishes a strong link with
convolutional neural networks.

\section{Convolutional Multilayer Kernels}\label{sec:scattering}
The convolutional multilayer kernel is a generalization of the hierarchical kernel
descriptors introduced in computer vision~\cite{bo2011,bo2010}. The kernel produces a
sequence of image representations that are built on top of each other in a
multilayer fashion. Each layer can be interpreted as a non-linear
transformation of the previous one with additional spatial invariance. We call these
layers \emph{image feature maps}\footnote{In the kernel literature, ``feature map'' denotes the mapping between data points and their representation in a reproducing kernel Hilbert space (RKHS)~\cite{shawe2004}. Here, feature maps refer to spatial maps representing local image characteristics at everly location, as usual in the neural network literature~\cite{lecun1998}.}, and formally define them as follows:
\begin{definition}
   An image feature map~$\varphi$ is a function~$\varphi : \Omega
   \to \HH$, where~$\Omega$ is a (usually discrete) subset of~$[0,1]^d$
   representing normalized ``coordinates'' in the image and~$\HH$ is a Hilbert space.
\end{definition}
For all practical examples in this paper, $\Omega$ is a two-dimensional grid
and corresponds to different locations in a two-dimensional image. In other
words, $\Omega$ is a set of pixel coordinates. Given~$\z$ in~$\Omega$, the
point~$\varphi(\z)$ represents some characteristics of the image at
location~$\z$, or in a neighborhood of~$\z$.
For instance, a color image of size~$m \times n$ with three
channels, red, green, and blue, may be represented by an initial feature
map~$\varphi_0: \Omega_0 \to \HH_0$, where~$\Omega_0$ is an~$m \times n$ regular grid, $\HH_0$ is the Euclidean space $\Real^3$, and $\varphi_0$
provides the color pixel values. With the multilayer scheme, non-trivial feature maps will be
obtained subsequently, which will encode more complex image characteristics.
With this terminology in hand, we now introduce the convolutional kernel, first, for a single layer.

\begin{definition}[\bfseries Convolutional Kernel with Single Layer] \label{def:scattering}
   Let us consider two images represented by two
   image feature maps, respectively~$\varphi$ and~$\varphi': \Omega \to \HH$,
   where~$\Omega$ is a set of pixel locations, and~$\HH$ is a Hilbert space.
   The one-layer convolutional kernel between~$\varphi$ and~$\varphi'$ is defined as 
  \begin{equation}
     K(\varphi,\varphi') \defin \sum_{\z \in \Omega} \sum_{\z' \in \Omega} \normH{\varphi(\z)}  \normH{\varphi'(\z')} e^{-\frac{1}{2\beta^2}\normE{\z-\z'}^2} e^{-\frac{1}{2\sigma^2} \normH{\tildephi(\z)-\tildephi'(\z')}^2}, \label{eq:kernel}
  \end{equation}
  where~$\beta$ and~$\sigma$ are smoothing parameters of Gaussian kernels,
  and~$\tildephi(\z) \!\defin\!  \left(1/\normH{\varphi(\z)}\right)\varphi(\z)$ if~$\varphi(\z) \neq 0$ and~$\tildephi(\z)=0$ otherwise.
  Similarly, $\tildephi'(\z')$ is a normalized version of~$\varphi'(\z')$.\footnote{When~$\Omega$ is not discrete, the notation~$\sum$ in~(\ref{eq:kernel}) should be replaced by the Lebesgue integral~$\int$ in the paper.}
\end{definition}
\vspace*{-0.1cm}
It is easy to show that the kernel~$K$ is positive definite (see Appendix~A). It consists of a sum of pairwise
comparisons between the image features~$\varphi(\z)$ and~$\varphi'(\z')$ computed at
all spatial locations~$\z$ and~$\z'$ in~$\Omega$. To be significant in the sum, a
comparison needs the corresponding $\z$ and~$\z'$ to be close 
in~$\Omega$, and the normalized features $\tildephi(\z)$ and
$\tildephi'(\z')$ to be close in the feature space~$\HH$. 
The parameters~$\beta$ and~$\sigma$ respectively control these two definitions
of ``closeness''. Indeed, when~$\beta$
is large, the kernel~$K$ is invariant to the positions~$\z$ and~$\z'$ but
when~$\beta$ is small, only features placed at the same location $\z=\z'$ are
compared to each other. Therefore, the role of~$\beta$ is to control how much the kernel
is locally shift-invariant. Next, we will show how to go beyond one single
layer, but before that, we present concrete examples of simple input
feature maps~$\varphi_0: \Omega_0 \to \HH_0$.
\vs
\paragraph{Gradient map.} Assume that $\HH_0\!=\!\Real^2$ and that
$\varphi_0(\z)$ provides the two-dimensional gradient of the image at
pixel~$\z$, which is often computed with first-order differences along each
dimension. Then, the quantity $\normHz{\varphi_0(\z)}$ is the gradient intensity, and~$\tildephi_0(\z)$ is its orientation,
which can be characterized by a particular angle---that is, there exists~$\theta$
in~$[0; 2\pi]$ such that $\tildephi_0(\z) = [\cos(\theta),\sin(\theta)]$. The
resulting kernel~$K$ is exactly the kernel descriptor introduced in
\cite{bo2011,bo2010} for natural image patches.

\vs
\paragraph{Patch map.} In that setting, $\varphi_0$ associates to a
location~$\z$ an image patch of size~$m \times m$ centered at~$\z$.
Then, the space~$\HH_0$ is simply~$\Real^{m \times m}$, and~$\tildephi_0(\z)$ is a \emph{contrast-normalized}
version of the patch, which is a useful transformation for visual recognition according
to classical findings in computer vision~\cite{jarrett2009}. When the
image is encoded with three color channels, patches are of size~$m \times m
\times 3$.

We now define the multilayer convolutional kernel,
generalizing some ideas of~\cite{bo2011}.

\begin{definition}[\bfseries Multilayer Convolutional Kernel]\label{def:multiscattering}
   Let us consider a set $\Omega_{\kmone} \subseteq [0,1]^d$ and a Hilbert space $\HH_{\kmone}$.
   We build a new set~$\Omega_k$ and a new Hilbert space~$\HH_k$ as follows:

   (i) choose a patch shape~$\PP_k$ defined as a bounded symmetric subset
   of~$[-1,1]^d$, and a set of coordinates~$\Omega_k$
   such that for all location~$\z_k$ in~$\Omega_k$, the patch~$\{\z_k\} + \PP_k$ is a subset of~$\Omega_{\kmone}$;\footnote{For two sets~$A$
   and~$B$, the Minkowski sum $A+B$ is defined as $\{ a + b : a \in A, b \in B\}$.\label{foot:minkowski}}
   In other words, each coordinate~$\z_k$ in~$\Omega_k$ corresponds to a valid patch in~$\Omega_{\kmone}$ centered at~$\z_k$.

   (ii) define the convolutional kernel~$K_k$ on the ``patch'' feature maps~$\PP_k \to
   \HH_{\kmone}$, by replacing in~(\ref{eq:kernel}):~$\Omega$ by~$\PP_k$,~$\HH$
   by~$\HH_{\kmone}$, and~$\sigma,\beta$ by appropriate smoothing
   parameters~$\sigma_k,\beta_k$. We denote by~$\HH_k$ the Hilbert space for
   which the positive definite kernel~$K_k$ is reproducing.

   An image represented by a feature map~$\varphi_{\kmone}: \Omega_{\kmone} \to
   \HH_{\kmone}$ at layer~$\kmone$ is now encoded in the~$k$-th layer as~$\varphi_k:
   \Omega_k \to \HH_k$, where for all~$\z_k$ in~$\Omega_k$,~$\varphi_{k}(\z_k)$ is the representation
   in~$\HH_k$ of the patch feature map~$\z \mapsto \varphi_{\kmone}(\z_k + \z)$ for~$\z$ in~$\PP_k$.
\end{definition}
Concretely, the kernel~$K_k$ between two patches of~$\varphi_{\kmone}$ and~$\varphi'_{\kmone}$ at respective locations~$\z_k$ and~$\z_k'$~is
\vspace*{-0.1cm}
\begin{equation}
   \sum_{\z \in \PP_k} \sum_{\z' \in \PP_k} \norm{\varphi_{\kmone}(\z_k+\z)}  \norm{\varphi_{\kmone}'(\z'_k + \z')} e^{-\frac{1}{2\beta_k^2}\normE{\z-\z'}^2} e^{-\frac{1}{2\sigma_k^2} \norm{\tildephi_{\kmone}(\z_k+\z)-\tildephi_{\kmone}'(\z_k'+\z')}^2}, \label{eq:kernelpatch}
\vspace*{-0.00cm}
\end{equation}
where~$\|.\|$ is the Hilbertian norm of~$\HH_{\kmone}$.
In Figure~\ref{subfig:hierarchy}, we illustrate the interactions
between the sets of coordinates~$\Omega_k$, patches~$\PP_k$, and feature spaces~$\HH_k$ across
layers. For two-dimensional grids, a typical patch
shape is a square, for example~$\PP := \{-1/n,0,1/n\} \times \{-1/n,0,1/n\}$ for a $3
\times 3$ patch in an image of size~$n \times n$. Information encoded in the~$k$-th layer differs from the~$(\kmone)$-th one in two aspects:
first, each point~$\varphi_k(\z_k)$ in layer~$k$ contains
information about several points from the $(\kmone)$-th layer and can possibly represent
larger patterns; second, the new feature map is more locally shift-invariant than the previous one due to the
term involving the parameter~$\beta_k$ in~(\ref{eq:kernelpatch}). 

\pgfdeclarelayer{bottom}  \pgfdeclarelayer{middle}\pgfdeclarelayer{top}
\pgfsetlayers{bottom,middle,top}  
\begin{figure}
   \subfigure[Hierarchy of image feature maps.]{\label{subfig:hierarchy}
      \hspace*{-0.4cm}
   \begin{tikzpicture}[scale=1,every node/.style={minimum size=1cm},on grid]
      \begin{pgfonlayer}{bottom}
         \begin{scope}[  
               yshift=0,every node/.append style={
               yslant=0.5,xslant=-1,rotate=-10},yslant=0.5,xslant=-1,rotate=-10
            ]
            \fill[white,fill opacity=0.9] (0,0) rectangle (3,3);
            \draw[step=2mm, gray!70] (0,0) grid (3,3);
            \draw[black] (0,0) rectangle (3,3);
            \draw[red!20,fill] (0.4,0.6) rectangle (0.6,0.8);
            \draw[red!80!black!100] (0.4,0.6) rectangle (0.6,0.8);
            \draw[blue!20,fill] (1,0.4) rectangle (1.8,1.2);
            \draw[blue!90] (1,0.4) rectangle (1.8,1.2);
            \draw[step=2mm, blue!70] (1,0.4) grid (1.8,1.2);
            \coordinate (a) at (0.5,0.7);
            \coordinate (B1) at (1,0.4);
            \coordinate (B2) at (1,1.2);
            \coordinate (B3) at (1.8,0.4);
            \coordinate (B4) at (1.8,1.2);
            \coordinate (B5) at (1.4,0.8);
            \coordinate (aa) at (1,0);
         \end{scope}
         \draw[-latex,thick] (1.8,0.1) node[right]{$\!\!\!\!\!\Omega_0$}to[out=180,in=-50] (aa);
         \draw[-latex,thick] (-0.3,0) node[left]{{\color{red!80!black!80} $\varphi_0(\z_0) \in \HH_0$}}to[out=0,in=270] (a);
         \draw[-latex,thick] (2.2,0.5) node[right]{{\color{blue!80!black!80} $\{\z_1\} + \PP_1 $}}to[out=180,in=50] (B5);
      \end{pgfonlayer}
      \begin{pgfonlayer}{middle}
         \begin{scope}[  
               yshift=50,every node/.append style={
               yslant=0.5,xslant=-1,rotate=-10},yslant=0.5,xslant=-1,rotate=-10
            ]
            \fill[white,fill opacity=.7] (0,0) rectangle (2.1,2.1);
            \draw[green!20,fill] (0,0.9) rectangle (0.9,1.8);
            \draw[step=3mm, gray!70] (0,0) grid (2.1,2.1);
            \draw[black] (0,0) rectangle (2.1,2.1);
            \draw[blue!20,fill] (0.9,0.3) rectangle (1.2,0.6);
            \draw[blue!90] (0.9,0.3) rectangle (1.2,0.6);
            \draw[green!70!black!100] (0,0.9) rectangle (0.9,1.8);
            \draw[step=3mm, green!70] (0,0.9) grid (0.9,1.8);
            \coordinate (b) at (1.05,0.45);
            \coordinate (c) at (0,1.5);
            \coordinate (A1) at (0.9,0.3);
            \coordinate (A2) at (0.9,0.6);
            \coordinate (A3) at (1.2,0.3);
            \coordinate (A4) at (1.2,0.6);
            \coordinate (D1) at (0,0.9);
            \coordinate (D2) at (0,1.8);
            \coordinate (D3) at (0.9,0.9);
            \coordinate (D4) at (0.9,1.8);
            \coordinate (D5) at (0.45,1.35);
         \end{scope}
      \draw[-latex,thick] (2.2,2.3) node[right]{{\color{blue!80!black!80}$\varphi_1(\z_1) \in \HH_1$}}to[out=180,in=50] (b);
      \draw[-latex,thick] (-1.8,2.4) node[left]{$\Omega_1\!\!\!\!\!\!$}to[out=0,in=180] (c);
      \draw[-latex,thick] (2.2,3.5) node[right]{{\color{green!60!black!100} $\{\z_2\} + \PP_2$}}to[out=180,in=50] (D5);
      \end{pgfonlayer}
      \begin{pgfonlayer}{bottom}
         \draw[thick,blue!70] (A1) -- (B1);
         \draw[thick,blue!70] (A2) -- (B2);
         \draw[thick,blue!70] (A3) -- (B3);
         \draw[thick,blue!70] (A4) -- (B4);
      \end{pgfonlayer}
      \begin{pgfonlayer}{top}
         \begin{scope}[  
               yshift=100,every node/.append style={
               yslant=0.5,xslant=-1,rotate=-10},yslant=0.5,xslant=-1,rotate=-10
            ]
            \fill[white,fill opacity=.7] (0,0) rectangle (1.6,1.6);
            \draw[step=4mm, gray!70] (0,0) grid (1.6,1.6);
            \draw[black] (0,0) rectangle (1.6,1.6);
            \draw[green!20,fill] (0,0.8) rectangle (0.4,1.2);
            \draw[green!70!black!100] (0,0.8) rectangle (0.4,1.2);
            \coordinate (CC1) at (0,1.6);
            \coordinate (CC2) at (0.2,1.0);
            \coordinate (C1) at (0,0.8);
            \coordinate (C2) at (0,1.2);
            \coordinate (C3) at (0.4,0.8);
            \coordinate (C4) at (0.4,1.2);
         \end{scope}
         \draw[-latex,thick] (-1.8,4.2) node[left]{$\Omega_2\!\!\!\!\!\!$}to[out=0,in=180] (CC1);
         \draw[-latex,thick] (2.2,4.8) node[right]{{\color{green!60!black!100}$\varphi_2(\z_2) \in \HH_2$}}to[out=180,in=50] (CC2);
      \end{pgfonlayer}
      \begin{pgfonlayer}{middle}
         \draw[thick,green!70!black!100] (C1) -- (D1);
         \draw[thick,green!70!black!100] (C2) -- (D2);
         \draw[thick,green!70!black!100] (C3) -- (D3);
         \draw[thick,green!70!black!100] (C4) -- (D4);
      \end{pgfonlayer}
   \end{tikzpicture}
   }
   \subfigure[Zoom between layer~$\kmone$ and~$k$ of the CKN.]{\label{subfig:convnet}
      \hspace*{-0.3cm}
      \begin{tikzpicture}[decoration={brace}][scale=1,every node/.style={minimum size=1cm},on grid]
         \begin{pgfonlayer}{bottom}
            \newcount\mycount
            \foreach \i in {0,1,2,3,4} {
               \mycount=\i
               \multiply\mycount by 3
               \begin{scope}[  
                     yshift=\mycount,every node/.append style={
                     yslant=0.5,xslant=-1,rotate=-10},yslant=0.5,xslant=-1,rotate=-10
                  ]
                  \coordinate (X\i) at (0.15,0.75);
                  \coordinate (G\i) at (1.5,0.45);
                  \coordinate (ZA\i) at (1.05,0);
                  \coordinate (ZB\i) at (1.35,0);
                  \coordinate (ZC\i) at (0.75,0);
                  \coordinate (AAA\i) at (0,0);
                  \coordinate (AAB\i) at (0,2.4);
                  \coordinate (AAC\i) at (2.4,0);
                  \coordinate (AAD\i) at (2.4,2.4);
                  \coordinate (AA\i) at (0.6,0);
                  \coordinate (AB\i) at (0.6,0.9);
                  \coordinate (AC\i) at (1.5,0);
                  \coordinate (AD\i) at (1.5,0.9);
                  \coordinate (EA\i) at (0,0.6);
                  \coordinate (EB\i) at (0,0.9);
                  \coordinate (EC\i) at (0.3,0.6);
                  \coordinate (ED\i) at (0.3,0.9);
                  \newcount\prevcount
                  \prevcount=\i
                  \advance\prevcount by -1
                  \ifnum\i>0
                  \draw[thick,blue!70] (AA\i) -- (AA\the\prevcount);
                  \draw[thick,blue!70] (AB\i) -- (AB\the\prevcount);
                  \draw[thick,blue!70] (AC\i) -- (AC\the\prevcount);
                  \draw[thick,blue!70] (AD\i) -- (AD\the\prevcount);
                  \draw[thick,red!70!black!100] (EA\i) -- (EA\the\prevcount);
                  \draw[thick,red!70!black!100] (EB\i) -- (EB\the\prevcount);
                  \draw[thick,red!70!black!100] (EC\i) -- (EC\the\prevcount);
                  \draw[thick,red!70!black!100] (ED\i) -- (ED\the\prevcount);
                  \draw[thick,black] (AAA\i) -- (AAA\the\prevcount);
                  \draw[thick,black] (AAB\i) -- (AAB\the\prevcount);
                  \draw[thick,black] (AAC\i) -- (AAC\the\prevcount);
                  \draw[thick,black] (AAD\i) -- (AAD\the\prevcount);
                  \fi
                  \fill[white,fill,opacity=.7] (0,0) rectangle (2.4,2.4);
                  \draw[step=3mm, gray!70] (0,0) grid (2.4,2.4);
                  \draw[black] (0,0) rectangle (2.4,2.4);
                  \draw[blue!20,fill] (0.6,0) rectangle (1.5,0.9);
                  \draw[blue!90] (0.6,0) rectangle (1.5,0.9);
                  \draw[step=3mm, blue!70] (0.6,0) grid (1.5,0.9);
                  \draw[red!20,fill] (0,0.6) rectangle (0.3,0.9);
                  \draw[red!80!black!100] (0,0.6) rectangle (0.3,0.9);
               \end{scope}
            }
            \draw[-latex,thick] (-1.9,0.9) node[below,xshift=1mm,yshift=2mm]{{\color{black} $\Omega'_{\kmone}$}}to[out=90,in=270] (AAB0);
            \draw[-latex,thick] (-0.8,0.2) node[left]{{\color{red!80!black!80} $\xi_{\kmone}(\z)$}}to[out=50,in=150] (X4);
            \draw[-latex,thick] (2.2,0.6) node[right]{{\color{blue!80!black!80} $\psi_{\kmone}(\z_{\kmone})$}}to[out=200,in=-90] (ZA4);
            \draw[-latex,thick,white] (2,0.25) node[right]{{\color{blue!80!black!80}\small (patch extraction)}}to[out=200,in=-90] (2.2,0.5);
            \draw[-latex,thick] (2.2,0.6) node[right]{}to[out=200,in=-90] (ZB4);
            \draw[-latex,thick] (2.2,0.6) node[right]{}to[out=200,in=-90] (ZC4);
            \draw[-latex,thick] (2.5,1.7) node[right]{{\color{blue!80!black!80} $\{\z_{\kmone}\} \!+\! \NN_{\kmone}$ }}to[out=190,in=0] (G4);
            \draw[-latex,thick] (2.2,2.4) node[right]{{\color{black!80} \small convolution}}to[out=190,in=0] (1.0,1.7);
            \draw[-latex,white] (2.2,2.1) node[right]{{\color{black!80} \small + non-linearity}}to[out=150,in=-90] (2.2,2.1);
         \end{pgfonlayer}
         \begin{pgfonlayer}{middle}
            \newcount\mycount
            \foreach \i in {0,1,2,3,4} {
               \mycount=\i
               \multiply\mycount by 3
               \advance\mycount by 55
               \begin{scope}[  
                     yshift=\mycount,every node/.append style={
                     yslant=0.5,xslant=-1,rotate=-10},yslant=0.5,xslant=-1,rotate=-10
                  ]
                  \coordinate (W\i) at (0.75,0.15);
                  \coordinate (BAA\i) at (0,0);
                  \coordinate (BAB\i) at (0,1.8);
                  \coordinate (BAC\i) at (1.8,0);
                  \coordinate (BAD\i) at (1.8,1.8);
                  \coordinate (BA\i) at (0.6,0);
                  \coordinate (BB\i) at (0.6,0.3);
                  \coordinate (BC\i) at (0.9,0);
                  \coordinate (BD\i) at (0.9,0.3);
                  \newcount\prevcount
                  \prevcount=\i
                  \advance\prevcount by -1
                  \ifnum\i>0
                  \draw[thick,blue!70] (BA\i) -- (BA\the\prevcount);
                  \draw[thick,blue!70] (BB\i) -- (BB\the\prevcount);
                  \draw[thick,blue!70] (BC\i) -- (BC\the\prevcount);
                  \draw[thick,blue!70] (BD\i) -- (BD\the\prevcount);
                  \draw[thick,black] (BAA\i) -- (BAA\the\prevcount);
                  \draw[thick,black] (BAB\i) -- (BAB\the\prevcount);
                  \draw[thick,black] (BAC\i) -- (BAC\the\prevcount);
                  \draw[thick,black] (BAD\i) -- (BAD\the\prevcount);
                  \fi
                  \fill[white,fill,opacity=.7] (0,0) rectangle (1.8,1.8);
                  \draw[step=3mm, gray!70] (0,0) grid (1.8,1.8);
                  \draw[black] (0,0) rectangle (1.8,1.8);
                  \draw[blue!20,fill] (0.6,0) rectangle (0.9,0.3);
                  \draw[blue!90] (0.6,0) rectangle (0.9,0.3);
               \end{scope}
            }
            \draw[decorate,decoration={brace,mirror,raise=1pt},line width=1pt] (BAC0) -- (BAC4) node[right,yshift=-2mm,xshift=1mm] {$p_k$};
            \draw[-latex,thick] (2.2,3.3) node[right]{{\color{blue!80!black!80} $\zeta_{k}(\z_{\kmone})$}}to[out=180,in=60] (W4);
            \draw[-latex,thick] (-1.7,2.7) node[below,yshift=3mm]{{\color{black} $\Omega_{\kmone}$}}to[out=90,in=180] (BAB0);
            \draw[-latex,thick] (2.2,4.3) node[right]{{\color{black!80} \small Gaussian filtering}}to[out=190,in=0] (1.3,4);
            \draw[-latex,thick,white] (2.2,4) node[right]{{\color{black!80} \small + downsampling}}to[out=190,in=0] (2.2,4);
            \draw[-latex,thick,white] (2.2,3.7) node[right]{{\color{black!80} \small = pooling}}to[out=190,in=0] (2.2,3.7);
         \end{pgfonlayer}
         \begin{pgfonlayer}{bottom}
            \draw[thick,blue!70] (BA0) -- (AA4);
            \draw[thick,blue!70] (BB0) -- (AB4);
            \draw[thick,blue!70] (BC0) -- (AC4);
            \draw[thick,blue!70] (BD0) -- (AD4);
         \end{pgfonlayer}
         \begin{pgfonlayer}{top}
            \foreach \i in {0,1,2,3,4} {
               \mycount=\i
               \multiply\mycount by 3
               \advance\mycount by 105
               \begin{scope}[  
                     yshift=\mycount,every node/.append style={
                     yslant=0.5,xslant=-1,rotate=-10},yslant=0.5,xslant=-1,rotate=-10
                  ]
                  \coordinate (H\i) at (1.25,0.75);
                  \coordinate (CAA\i) at (0,0);
                  \coordinate (CAB\i) at (0,1.5);
                  \coordinate (CAC\i) at (1.5,0);
                  \coordinate (CAD\i) at (1.5,1.5);
                  \coordinate (CA\i) at (1.0,0.5);
                  \coordinate (CB\i) at (1.0,1.0);
                  \coordinate (CC\i) at (1.5,0.5);
                  \coordinate (CD\i) at (1.5,1.0);
                  \newcount\prevcount
                  \prevcount=\i
                  \advance\prevcount by -1
                  \ifnum\i>0
                  \draw[thick,black] (CAA\i) -- (CAA\the\prevcount);
                  \draw[thick,black] (CAB\i) -- (CAB\the\prevcount);
                  \draw[thick,black] (CAC\i) -- (CAC\the\prevcount);
                  \draw[thick,black] (CAD\i) -- (CAD\the\prevcount);
                  \draw[thick,green!70!black!100] (CA\i) -- (CA\the\prevcount);
                  \draw[thick,green!70!black!100] (CB\i) -- (CB\the\prevcount);
                  \draw[thick,green!70!black!100] (CC\i) -- (CC\the\prevcount);
                  \draw[thick,green!70!black!100] (CD\i) -- (CD\the\prevcount);
                  \fi
                  \fill[white,fill,opacity=.7] (0,0) rectangle (1.5,1.5);
                  \draw[step=5mm, gray!70] (0,0) grid (1.5,1.5);
                  \draw[black] (0,0) rectangle (1.5,1.5);
                  \draw[green!20,fill] (1.0,0.5) rectangle (1.5,1.0);
                  \draw[green!70!black!100] (1.0,0.5) rectangle (1.5,1.0);
               \end{scope}
            }
            \draw[-latex,thick] (-1.6,4.2) node[below,yshift=3mm]{{\color{black} $\Omega'_{k}$}}to[out=90,in=180] (CAB0);
            \draw[-latex,thick] (2.2,5.2) node[right]{{\color{green!60!black!100} $\xi_{k}(\z)$}}to[out=180,in=60] (H4);
         \end{pgfonlayer}
         \begin{pgfonlayer}{middle}
            \draw[thick,black!50] (CAA0) -- (BAA4);
            \draw[thick,black!50] (CAB0) -- (BAB4);
            \draw[thick,black!50] (CAC0) -- (BAC4);
            \draw[thick,black!50] (CAD0) -- (BAD4);
         \end{pgfonlayer}
      \end{tikzpicture}
      \label{subfig:cnn}
   }
   \vspace*{-0.3cm}
   \caption{Left: concrete representation of the successive layers for the multilayer convolutional kernel. Right: one layer of the convolutional neural network that approximates the kernel.
   }\label{fig:sketch}
\end{figure}
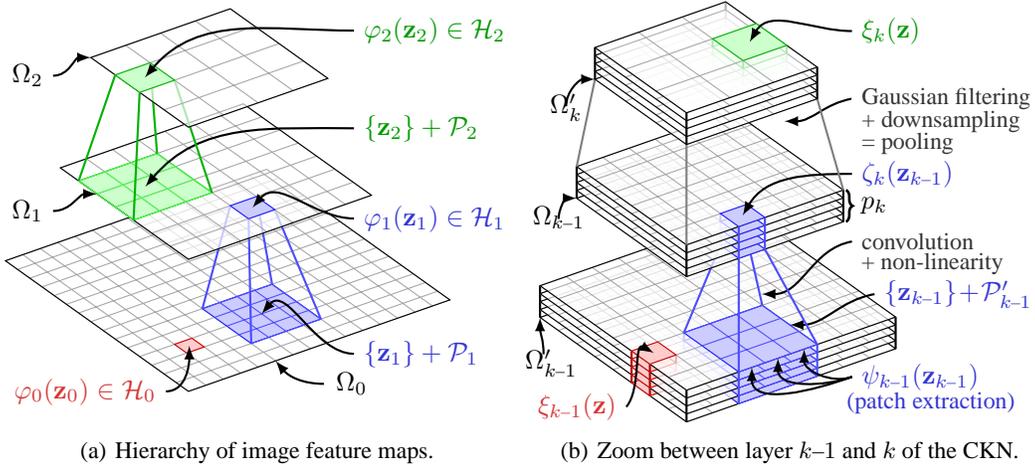

The multilayer convolutional kernel slightly differs from the hierarchical kernel descriptors
of~\cite{bo2011} but exploits similar ideas. Bo et al.~\cite{bo2011} define
indeed several ad hoc kernels for representing local information in images,
such as gradient, color, or shape. These kernels are close to the one
defined in~(\ref{eq:kernel}) but with a few variations. Some of them do not
use normalized features~$\tildephi(\z)$, and these kernels use different
weighting strategies for the summands of~(\ref{eq:kernel}) that are specialized
to the image modality, \eg, color, or gradient, whereas we use the same
weight~$\normH{\varphi(\z)}  \normH{\varphi'(\z')}$ for all kernels. The generic
formulation~(\ref{eq:kernel}) that we propose may be useful per
se, but our main contribution comes in the next section, where we use the  
kernel as a new tool for learning convolutional neural networks.

\section{Training Invariant Convolutional Kernel Networks}\label{sec:approx}
Generic schemes have been proposed for approximating a non-linear kernel with a
linear one, such as the Nystr\"om method and its
variants~\cite{bo2009,williams2001}, or random sampling techniques in the
Fourier domain for shift-invariant kernels~\cite{rahimi2007}.  In the context
of convolutional multilayer kernels, such an approximation is critical because
computing the full kernel matrix on a database of images is computationally
infeasible, even for a moderate number of images ($\approx 10\,000$) and
moderate number of layers. For this reason, Bo et al.~\cite{bo2011}
use the Nystr\"om method for their hierarchical kernel descriptors. 

In this section, we show that when the coordinate sets~$\Omega_k$ are 
two-dimensional regular grids, a natural approximation for the multilayer convolutional kernel consists of a sequence of
spatial convolutions with learned filters, pointwise non-linearities, and pooling
operations, as illustrated in Figure~\ref{subfig:convnet}. 
More precisely, our scheme approximates the kernel map of~$K$ defined
in~(\ref{eq:kernel}) at layer~$k$ by finite-dimensional spatial maps~$\xi_k: \Omega'_k \to \Real^{p_k}$, where~$\Omega'_k$ is a set of coordinates related to~$\Omega_k$,
and~$p_k$ is a positive integer controlling the quality of the approximation. 
Consider indeed two images represented at layer~$k$ by image
feature maps $\varphi_k$ and~$\varphi_k'$, respectively. Then,
\begin{itemize}[leftmargin=0.7cm]
   \item[{\bfseries (A)}] the corresponding maps~$\xi_k$ and~$\xi'_k$ are learned such that
      $K(\varphi_{\kmone},\varphi_{\kmone}') \approx \langle \xi_k, \xi_k' \rangle$, where~$\langle.,.\rangle$ is
      the Euclidean inner-product acting as if~$\xi_k$ and~$\xi_k'$ were vectors in~$\Real^{|\Omega_k'| p_k}$;
   \item[{\bfseries (B)}] the set~$\Omega'_k$ is linked to~$\Omega_k$ by the
      relation~$\Omega_k'\!=\!\Omega_k + \PP_k'$ where~$\PP_k'$ is a patch
      shape, and the quantities~$\varphi_k(\z_k)$ in~$\HH_k$ admit finite-dimensional 
      approximations~$\psi_k(\z_k)$ in~$\Real^{|\PP_k'|p_k}$; as
      illustrated in Figure~\ref{subfig:convnet}, $\psi_k(\z_k)$ is a patch
      from~$\xi_k$ centered at location~$\z_k$ with shape~$\PP_k'$;
   \item[{\bfseries (C)}] an activation map~$\zeta_{k}: \Omega_{\kmone} \mapsto \Real^{p_{k}}$ is computed from~$\xi_{\kmone}$ by 
      convolution with~$p_k$ filters followed by a non-linearity. The subsequent map~$\xi_k$ is obtained from~$\zeta_k$ by a
      pooling operation. 
\end{itemize}
We call this approximation scheme a convolutional kernel network
(CKN). In comparison to CNNs, our approach enjoys similar benefits such as efficient prediction at test
time, and involves the same set of hyper-parameters: number of layers, numbers
of filters~$p_k$ at layer~$k$, shape~$\PP_k'$ of the filters, sizes of the feature maps.
The other parameters~$\beta_k, \sigma_k$ can be automatically chosen, as
discussed later. Training a CKN can be argued to be as simple as
training a CNN in an unsupervised manner~\cite{ranzato2007} since we will show that
the main difference is in the cost function that is optimized.

\subsection{Fast Approximation of the Gaussian Kernel}\label{subsec:approx_gaussian}
A key component of our formulation is the Gaussian kernel. We start by
approximating it by a linear operation with learned filters followed by a
pointwise non-linearity.  Our starting point is the next lemma, which can be
obtained after a simple calculation.
\begin{lemma}[\bfseries Linear expansion of the Gaussian Kernel]
   For all~$\x$ and~$\x'$ in~$\Real^m$, and~$\sigma > 0$, 
   \begin{equation}
      e^{-\frac{1}{2\sigma^2}\|\x-\x'\|_2^2} = \left(\frac{2}{\pi \sigma^2}\right)^{\frac{m}{2}} \int_{\w \in \Real^m} e^{-\frac{1}{\sigma^2}\|\x-\w\|_2^2}e^{-\frac{1}{\sigma^2}\|\x'-\w\|_2^2} d\w.\label{eq:rbf}
   \end{equation}
\end{lemma}
The lemma gives us a mapping of any~$\x$ in~$\Real^m$ to the function $\w
\mapsto \sqrt{C} e^{-(1/\sigma^2)\|\x-\w\|_2^2}$ in $L^2(\Real^m)$, where the
kernel is linear, and $C$ is the constant in front of the integral. 
To obtain a finite-dimensional representation, we need
to approximate the integral with a weighted finite sum, which is a
classical problem arising in statistics~(see \cite{wahba} and chapter 8 of~\cite{bottou2007}). 
Then, we consider two different~cases.
\vspace*{-0.2cm}
\paragraph{Small dimension, $m \leq 2$.}
When the data lives in a compact set of~$\Real^m$, the
integral in~(\ref{eq:rbf}) can be approximated by uniform sampling over a large
enough set. We choose such a strategy for two types of kernels from Eq.~(\ref{eq:kernel}):
(i) the spatial kernels~$e^{-\left(\frac{1}{2\beta^2}\right)\left\|\z-\z'\right\|_2^2}$;
(ii) the terms $e^{-\left(\frac{1}{2\sigma^2}\right)\left\|\tildephi(\z)-\tildephi'(\z')\right\|_\HH^2}$ when
$\varphi$ is the ``gradient map'' presented in Section~\ref{sec:scattering}. In
the latter case, $\HH = \Real^2$ and $\tildephi(\z)$ is the gradient orientation.
We typically sample a few orientations as explained in Section~\ref{sec:exp}.

\begin{figure}[hbtp]
   \centering
   \begin{tikzpicture}[scale=2]
      \draw[->,very thick,black] (-1,0) -- (1,0) node[right] {$u$};
      \draw[->,very thick,black] (0,-0.1) -- (0,1) node[right,yshift=-2mm] {$f(u)$};
      \draw[scale=1,domain=-1:1,smooth,variable=\x,very thick,blue] plot ({\x},{exp(2*(\x-1))});
      \draw[scale=1,domain=-1:1,smooth,variable=\x,very thick,blue] plot ({\x},{exp(4*(\x-1))}) node[left,xshift=-3cm,yshift=-0.5cm] {$f(u)=e^{(2/\sigma^2)(u-1)}$};
      \draw[scale=1,domain=-1:1,smooth,variable=\x,very thick,blue] plot ({\x},{exp(6*(\x-1))});
      \draw[scale=1,domain=-1:1,smooth,variable=\x,very thick,dashed,red] plot ({\x},{max(\x,0)}) node[left,xshift=-3.29cm,yshift=-1cm] {$f(u)=\max(u,0)$};
      \draw (0,0) node[anchor=north,xshift=2mm] {0};
      \draw (1,0) node[anchor=north] {1};
      \draw (-1,0) node[anchor=north] {-1};
      \draw[very thick] (-1,0.05) -- (-1,-0.05);
      \draw[very thick] (1,0.05) -- (1,-0.05);
   \end{tikzpicture}
   \vspace*{-0.3cm}
   \caption{In dotted red, we plot the ``rectified linear unit'' function $u \mapsto \max(u,0)$. In blue, we plot non-linear functions of our network for typical values of~$\sigma$ that we use in our experiments.}\label{fig:relu}
\end{figure}
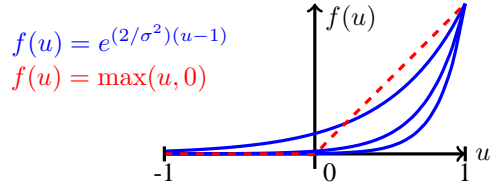

\vs
\paragraph{Higher dimensions.}
To prevent the curse of dimensionality, we learn to
approximate the kernel on training data, which is intrinsically
low-dimensional. We optimize importance weights~$\etab=[\eta_l]_{l=1}^p$
in~$\Real_+^p$ and sampling points $\W = [\w_l]_{l=1}^p$ in~$\Real^{m \times
p}$ on~$n$ training pairs~$(\x_i,\y_i)_{i=1,\ldots,n}$ in~$\Real^m \times
\Real^m$:
\begin{equation}
   \min_{\etab \in \Real_+^p, \W \in \Real^{m \times p}} \bigg[ \frac{1}{n} \sum_{i=1}^n \Big(e^{-\frac{1}{2\sigma^2}\|\x_i-\y_i\|_2^2} - \sum_{l=1}^p \eta_l e^{-\frac{1}{\sigma^2}\|\x_i-\w_l\|_2^2} e^{-\frac{1}{\sigma^2}\|\y_i-\w_l\|_2^2}\Big)^2   \bigg]. \label{eq:opt}
\end{equation}
Interestingly, we may already draw some links with neural networks. When
applied to unit-norm vectors $\x_i$ and~$\y_i$,
problem~(\ref{eq:opt}) produces sampling points~$\w_l$ whose norm is close to
one. After learning, a new unit-norm point~$\x$ in~$\Real^m$ is mapped to the
vector $[\sqrt{\eta_l} e^{-({1}/{\sigma^2})\|\x-\w_l\|_2^2}]_{l=1}^p$ in~$\Real^p$,
which may be written as $[f(\w_l^\top \x)]_{l=1}^p$, assuming that the norm of~$\w_l$ is always one, where $f$ is the
function $u \mapsto e^{(2/\sigma^2)(u -1)}$ for
$u=\w_l^\top\x$ in~$[-1, 1]$.  
Therefore, the finite-dimensional representation of~$\x$ only involves a linear
operation followed by a non-linearity, as in typical neural networks.  In
Figure~\ref{fig:relu}, we show that the shape of~$f$ resembles the ``rectified
linear unit'' function~\cite{wan2013}.
\pgfsetlayers{main}

\vspace*{-0.3cm}
\subsection{Approximating the Multilayer Convolutional Kernel}

We have now all the tools in hand to build our convolutional kernel network.
We start by making assumptions on the input data, and then present the learning scheme and its approximation principles.

\vs
\paragraph{The zeroth layer.} We assume that the input data is a
finite-dimensional map $\xi_0: \Omega_0' \to \Real^{p_0}$, and that $\varphi_0:
\Omega_0 \to \HH_0$ ``extracts'' patches from~$\xi_0$. Formally, there
exists a patch shape~$\PP_0'$ such that $\Omega_0' = \Omega_0 + \PP_0'$, $\HH_0
= \Real^{p_0|\PP_0'|}$, and for all $\z_0$ in~$\Omega_0$, $\varphi_0(\z_0)$ is
a patch of $\xi_0$ centered at~$\z_0$. 
Then, property {\bfseries (B)} described at the beginning of
Section~\ref{sec:approx} is satisfied for $k\!=\!0$ by choosing
$\psi_0\!=\!\varphi_0$. The examples of input feature maps given earlier
satisfy this finite-dimensional assumption: for the gradient map, $\xi_0$ is
the gradient of the image along each direction, with $p_0=2$, $\PP_0'=\{ 0\}$
is
a $1 \!\times \!1$ patch, $\Omega_0\!=\!\Omega_0'$, and $\varphi_0\!=\!\xi_0$;
for the patch map, $\xi_0$ is the input image, say with $p_0\!=\!3$ for RGB data.

\vs
\paragraph{The convolutional kernel network.}
The zeroth layer being characterized, we present in Algorithms~\ref{alg:ckn}
and~\ref{alg:ckn2} the subsequent layers and how to learn their parameters 
in a feedforward manner. It is interesting to note that the input parameters of
the algorithm are exactly the same as a CNN---that is, number of layers and filters,
sizes of the patches and feature maps (obtained here via the subsampling factor).
Ultimately, CNNs and CKNs only differ in the cost function that is
optimized for learning the filters and in the choice of non-linearities.  As we
show next, there exists a link between the parameters of a CKN and those of
a convolutional multilayer kernel.

\begin{algorithm}
   \caption{Convolutional kernel network - learning the parameters of the $k$-th layer.}\label{alg:ckn}
   \begin{algorithmic}[1]
      \INPUT $\xi^1_{\kmone}, \xi^2_{\kmone},\ldots: \Omega'_{\kmone} \to \Real^{p_{\kmone}}$ (sequence of $(\kmone)$-th maps obtained from training images); $\PP_{\kmone}'$ (patch shape); $p_k$ (number of filters); $n$ (number of training pairs);
      \STATE extract at random $n$ pairs $(\x_i,\y_i)$ of patches with shape $\PP_{\kmone}'$ from the maps~$\xi_{\kmone}^1,\xi_{\kmone}^2,\ldots$;
      \STATE if not provided by the user, set $\sigma_k$ to the~$0.1$ quantile of the data~($\|\x_i-\y_i\|_2)_{i=1}^n$;
      \STATE {\bfseries unsupervised learning:} optimize~(\ref{eq:opt}) to obtain the filters~$\W_k$ in~$\Real^{|\PP_{\kmone}'|p_{\kmone} \times p_k}$ and $\etab_k$ in~$\Real^{p_k}$;
      \OUTPUT $\W_k$, $\etab_k$, and $\sigma_k$ (smoothing parameter);
   \end{algorithmic}
\end{algorithm}
\begin{algorithm}
   \caption{Convolutional kernel network - computing the $k$-th map form the~$(\kmone)$-th one.}\label{alg:ckn2}
   \begin{algorithmic}[1]
      \INPUT $\xi_{\kmone}: \Omega'_{\kmone} \!\to\! \Real^{p_{\kmone}}$ (input map); $\PP_{\kmone}'$ (patch shape); $\gamma_k \!\geq \!1$ (subsampling factor); $p_k$ (number of filters); $\sigma_k$ (smoothing parameter); $\W_k=[\w_{kl}]_{l=1}^{p_k}$ and~$\etab_k=[\eta_{kl}]_{l=1}^{p_k}$ (layer parameters);
      \STATE {\bfseries convolution and non-linearity:} define the activation map $\zeta_k: \Omega_{\kmone} \to \Real^{p_k}$ as
      \begin{equation}
         \vsb
         \zeta_k: \z \mapsto \|\psi_{\kmone}(\z)\|_2 \left[\sqrt{\eta_{kl}} e^{-\frac{1}{\sigma_k^2}\left\|\tildepsi_{\kmone}(\z)-\w_{kl}\right\|_2^2}\right]_{l=1}^{p_k}, \label{eq:zeta}
      \end{equation}
      where~$\psi_{\kmone}(\z)$ is a vector representing a patch from~$\xi_{\kmone}$ centered at~$\z$ with shape $\PP_{\kmone}'$, and the vector $\tildepsi_{\kmone}(\z)$ is an $\ell_2$-normalized version of~$\psi_{\kmone}(\z)$. This operation can be interpreted as a spatial convolution of the map~$\xi_{\kmone}$ with the filters~$\w_{kl}$ followed by pointwise non-linearities;
      \STATE set~$\beta_k$ to be~$\gamma_k$ times the spacing between two pixels in~$\Omega_{\kmone}$;
      \STATE {\bfseries feature pooling:}
      $\Omega'_k$ is obtained by subsampling~$\Omega_{\kmone}$ by a factor~$\gamma_k$ and we define a    
      new map $\xi_{k}: \Omega'_k \to \Real^{p_k}$ obtained from~$\zeta_k$ by linear pooling with Gaussian weights:
      \begin{equation}
         \vsb
         \xi_k: \z \mapsto \sqrt{{2}/\pi}\sum_{\u \in \Omega_{\kmone}}  e^{-\frac{1}{\beta_k^2}\left\|\u - \z\right\|_2^2} \zeta_k(\u). \label{eq:xi}
      \end{equation}
      \OUTPUT $\xi_{k} : \Omega'_k \to \Real^{p_k}$ (new map);
   \end{algorithmic}
\end{algorithm}

\paragraph{Approximation principles.}
\vspace*{-0.75cm}
We proceed recursively to show that the kernel approximation
property~{\bfseries (A)}
is satisfied; we assume that~{\bfseries (B)}
holds at layer~$\kmone$, and then, we show that {\bfseries (A)} and
{\bfseries (B)} also hold at layer~$k$.  This is sufficient for our 
purpose since we have previously assumed~{\bfseries (B)} for the zeroth layer.  
Given two images feature maps~$\varphi_{\kmone}$ and~$\varphi_{\kmone}'$, we 
start by approximating $K(\varphi_{\kmone},\varphi_{\kmone}')$ by replacing
$\varphi_{\kmone}(\z)$ and $\varphi_{\kmone}'(\z')$ by their finite-dimensional
approximations provided by~{\bfseries (B)}: 
\begin{equation}
   K(\varphi_{\kmone},\varphi_{\kmone}') \approx 
   \sum_{\z,\z' \in \Omega_{\kmone}} \normE{\psi_{\kmone}(\z)}  \normE{\psi_{\kmone}'(\z')} e^{-\frac{1}{2\beta_{k}^2}\normE{\z-\z'}^2} e^{-\frac{1}{2\sigma_k^2} \normE{\tildepsi_{\kmone}(\z)-\tildepsi_{\kmone}'(\z')}^2}.
\end{equation}
Then, we use the finite-dimensional approximation of the Gaussian kernel
involving~$\sigma_k$ and
\begin{equation}
         \vsb
   K(\varphi_{\kmone},\varphi_{\kmone}') \approx 
   \sum_{\z, \z' \in \Omega_{\kmone}} \zeta_k(\z)^\top \zeta_k'(\z') e^{-\frac{1}{2\beta_k^2}\normE{\z-\z'}^2},
\end{equation}
where~$\zeta_k$ is defined in~(\ref{eq:zeta}) and $\zeta_k'$ is defined
similarly by replacing~$\tildepsi$ by~$\tildepsi'$.  Finally, we approximate
the remaining Gaussian kernel by uniform sampling on $\Omega_{k}'$,
following Section~\ref{subsec:approx_gaussian}.
After exchanging sums and grouping appropriate terms together, we obtain the new approximation 
\begin{equation}
   K(\varphi_{\kmone},\varphi_{\kmone}') \approx \frac{2}{\pi} \sum_{\u \in \Omega_{k}'} \bigg( \sum_{\z \in \Omega_{\kmone}} e^{-\frac{1}{\beta_k^2}\normE{\z-\u}^2}\zeta_k(\z) \bigg)^\top \bigg( \sum_{\z' \in \Omega_{\kmone}}  e^{-\frac{1}{\beta_k^2}\normE{\z'-\u}^2} \zeta_k'(\z')  \bigg), \label{eq:approx}
\end{equation}
where the constant~$2/\pi$ comes from the multiplication of the constant
$2/(\pi\beta_k^2)$ from~(\ref{eq:rbf}) and the weight $\beta_k^2$ of uniform sampling
orresponding to the square of the distance between two pixels
of~$\Omega_k'$.\footnote{The choice of~$\beta_k$ in Algorithm~\ref{alg:ckn2} is
driven by signal processing principles. The feature pooling step can indeed be
interpreted as a downsampling operation that reduces the resolution of the map
from~$\Omega_{\kmone}$ to~$\Omega_k$ by using a Gaussian anti-aliasing filter,
whose role is to reduce frequencies
above the Nyquist limit.} As a result, the right-hand side is exactly $\langle
\xi_{k}, \xi_k' \rangle$, where~$\xi_k$ is defined in~(\ref{eq:xi}), giving us
property~{\bfseries (A)}. It remains to show that property~{\bfseries (B)} also
holds, specifically that the quantity~(\ref{eq:kernelpatch}) can be
approximated by the Euclidean inner-product~$\langle\psi_{k}(\z_k),
\psi'_k(\z_k')\rangle$ with the patches~$\psi_k(\z_k)$ and~$\psi'_k(\z_k')$ of shape~$\PP_k'$; we assume for that purpose that~$\PP_k'$ is a
subsampled version of the patch shape~$\PP_k$ by a factor~$\gamma_k$.

We remark that the kernel~(\ref{eq:kernelpatch}) is the same
as~(\ref{eq:kernel}) applied to layer~$\kmone$ by replacing~$\Omega_{\kmone}$
by~$\{\z_k\}+\PP_k$. By doing the same substitution in~(\ref{eq:approx}), we
immediately obtain an approximation of~(\ref{eq:kernelpatch}). Then, all
Gaussian terms are negligible for all~$\u$ and~$\z$ that are far from each
other---say when $\|\u-\z\|_2 \geq 2\beta_k$. Thus, we may replace the
sums~$\sum_{\u \in \Omega_k'}\sum_{\z,\z' \in \{\z_k\}+\PP_k}$ by~$\sum_{\u \in \{\z_k\}+\PP_k'}\sum_{\z,\z' \in \Omega_{\kmone}}$,
which has the same set of ``non-negligible'' terms. This yields exactly the
approximation $\langle\psi_{k}(\z_k), \psi'_k(\z_k')\rangle$.

\vs
\paragraph{Optimization.} 
Regarding problem~(\ref{eq:opt}), stochastic gradient descent
(SGD) may be used since a potentially infinite amount of training
data is available. However, we have preferred to use
L-BFGS-B~\cite{byrd1995} on $300\,000$ pairs of randomly selected training data
points, and initialize~$\W$ with the K-means algorithm. L-BFGS-B
is a parameter-free state-of-the-art batch method, which is not as
fast as SGD but much easier to use. We always run the L-BFGS-B algorithm for~$4\,000$ iterations, which seems to ensure
convergence to a stationary point. Our goal is to demonstrate the preliminary
performance of a new type of convolutional network, and we leave as future work
any speed improvement. 
\vsb

\section{Experiments}\label{sec:exp}
\vsb
We now present experiments that were performed using Matlab and an L-BFGS-B
solver~\cite{byrd1995} interfaced by Stephen Becker. Each image is represented by the last map~$\xi^k$ of the CKN, which is used in a linear SVM implemented in the
software package LibLinear~\cite{fan2}. 
These representations are centered, rescaled to have unit~$\ell_2$-norm on
average, and the regularization parameter of the SVM is always selected on a validation
set or by $5$-fold cross-validation in the range~$2^i$,
$i=-15\ldots,15$.

The patches~$\PP_k'$ are typically small; we tried the sizes~$m \times m$
with~$m=3,4,5$ for the first layer, and~$m=2,3$ for the upper ones. The number
of filters~$p_k$ in our experiments is in the set~$\{50,100,200,400,800\}$. The
downsampling factor~$\gamma_k$ is always chosen to be~$2$ between two
consecutive layers, whereas the last layer is downsampled to produce final
maps~$\xi_k$ of a small size---say, $5 \times 5$ or~$4 \times 4$. 
For the gradient map~$\varphi_0$, we approximate the Gaussian kernel
$e^{(1/\sigma_1^2)\|\varphi_0(\z)-\varphi_0'(\z')\|_{\HH_0}}$ by uniformly
sampling~$p_1=12$ orientations, setting~$\sigma_1=2\pi/p_1$.
Finally, we also use a small offset~$\varepsilon$ to prevent numerical
instabilities in the normalization steps~$\tildepsi(\z)= \psi(\z)/\max(\|\psi(\z)\|_2,\varepsilon)$.

\begin{figure}[b!]
   \centering
\vs
\includegraphics[width=12.5cm]{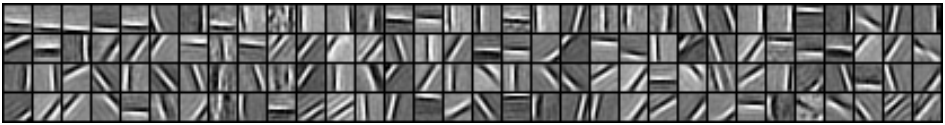}
\vs
\caption{Filters obtained by the first layer of the convolutional kernel network on natural images.}\label{fig:gabors}
\end{figure}

\vsb
\subsection{Discovering the Structure of Natural Image Patches}\label{sec:gabors}
\vsb
Unsupervised learning was first used for discovering the underlying structure
of natural image patches by Olshausen and Field~\cite{olshausen}.  Without
making any a priori assumption about the data except a parsimony principle, the
method is able to produce small prototypes that resemble Gabor wavelets---that
is, spatially localized oriented basis functions. The results were found
impressive by the scientific community and their work received substantial
attention. It is also known that such results can also be achieved with
CNNs~\cite{ranzato2007}. We show in this section that this is also the case for
convolutional kernel networks, even though they are not explicitly trained to reconstruct
data.

Following~\cite{olshausen}, we randomly select a database of~$300\,000$
whitened natural image patches of size~$12 \times 12$ and learn~$p=256$
filters~$\W$ using the formulation~(\ref{eq:opt}). We initialize~$\W$
with Gaussian random noise without performing the K-means step, in order to ensure that
the output we obtain is not an artifact of the initialization.
In Figure~\ref{fig:gabors}, we display the filters associated to the
top-$128$ largest weights~$\eta_l$. Among the~$256$ filters,~$197$ exhibit
interpretable Gabor-like structures and the rest was less interpretable.
To the best of our knowledge, this is the first time that the explicit kernel
map of the Gaussian kernel for whitened natural image patches is shown to be
related to Gabor wavelets.

\subsection{Digit Classification on MNIST}
The MNIST dataset~\cite{lecun1998} consists of $60\,000$ images of handwritten
digits for training and~$10\,000$ for testing. We use two types of initial
maps in our networks: the ``patch map'', denoted by~CNK-PM and the
``gradient map'', denoted by CNK-GM. We follow the evaluation methodology
of~\cite{ranzato2007} for comparison
when varying the training set size. 
We select the regularization parameter of the SVM by~$5$-fold cross validation when the
training size is smaller than~$20\,000$, or otherwise, we keep~$10\,0000$
examples from the training set for validation.  We report in
Table~\ref{table:mnist} the results obtained for four simple architectures.
CKN-GM1 is the simplest one: its second layer uses $3 \times 3$ patches and
only~$p_2=50$ filters, resulting in a network with~$5\,400$ parameters. Yet, it
achieves an outstanding performance of~$0.58\%$ error on the full dataset.
The best performing, CKN-GM2, is similar to CKN-GM1 but uses $p_2=400$ filters.
When working with raw patches, two layers (CKN-PM2) gives better results than
one layer.  More details about the network architectures are provided in the
supplementary material. In general, our method achieves 
a state-of-the-art accuracy for this task since lower error rates have only
been reported by using data augmentation~\cite{ciresan2012}.

\begin{table}
   \centering
   \renewcommand\tabcolsep{0.16cm}
   \footnotesize
\begin{tabular}{ | c | c |c| c || c | c | c | c || c |c |c|}
\hline
Tr. & CNN & Scat-1 & Scat-2 & CKN-GM1 & CKN-GM2 & CKN-PM1 & CKN-PM2 & \multirow{2}{*}{\cite{zeiler2013}} & \multirow{2}{*}{\cite{goodfellow2013}} & \multirow{2}{*}{\cite{jarrett2009}}  \\
size & \cite{ranzato2007} & \cite{bruna2013} & \cite{bruna2013} & ($12/50$) & ($12/400$) & ($200$) & ($50/200$) &  & & \\
\hline
$300$   & $7.18$ & $4.7$ & $5.6$ & $4.39$ & $4.24$ & $5.98$ & {\bfseries 4.15} & \multicolumn{3}{c|}{NA}\\
$1K$    & $3.21$ & $2.3$ & $2.6$ &   $2.60$ & {\bfseries 2.05} & $3.23$ & $2.76$ & \multicolumn{3}{c|}{NA}\\
$2K$    & $2.53$ & {\bfseries 1.3} & $1.8$ & $1.85$ & $1.51$ & $1.97$ & $2.28$ &  \multicolumn{3}{c|}{NA}\\
$5K$    & $1.52$ & {\bfseries 1.03} & $1.4$ &  $1.41$ & $1.21$ & $1.41$ & $1.56$ &  \multicolumn{3}{c|}{NA}\\
$10K$   & $0.85$ & {\bfseries 0.88} & $1$ & $1.17$ & {\bfseries 0.88} & $1.18$ & $1.10$ &  \multicolumn{3}{c|}{NA}\\
$20K$   & $0.76$ & $0.79$ & {\bfseries 0.58} & $0.89$ & $0.60$ & $0.83$ & $0.77$ &  \multicolumn{3}{c|}{NA}\\
$40K$   & $0.65$ & $0.74$ & $0.53$ & $0.68$ & {\bfseries 0.51} & $0.64$ & $0.58$ &  \multicolumn{3}{c|}{NA}\\
\hline
$60K$   & $0.53$ & $0.70$ & $0.4$ & $0.58$ & {\bfseries 0.39} & $0.63$ & $0.53$ & $0.47$ & $0.45$ & 0.53 \\
\hline
\end{tabular}
\caption{Test error in~$\%$ for various approaches on the MNIST dataset without data augmentation. The numbers in parentheses represent the size~$p_1$ and~$p_2$ of the feature maps at each layer.}\label{table:mnist}
\vs
\vs
\end{table}

\vsb
\subsection{Visual Recognition on CIFAR-10 and STL-10}
We now move to the more challenging datasets
CIFAR-10~\cite{krizhevsky2009} and STL-10~\cite{coates2011}. We select the
best architectures on a validation set of~$10\,000$ examples from the training
set for CIFAR-10, and by~$5$-fold cross-validation on STL-10. We report in Table~\ref{table:cifar} results for CKN-GM,
defined in the previous section, without exploiting color information, and
CKN-PM when working on raw RGB patches whose
mean color is subtracted. The best selected models have always two layers,
with~$800$ filters for the top layer.  Since CKN-PM and CKN-GM exploit a
different information, we also report a combination of such two models, CKN-CO,
by concatenating normalized image representations together. The standard deviations
for STL-10 was always below $0.7\%$.
Our approach appears to be competitive with the state of the
art, especially on STL-10 where only one method does better than ours, despite
the fact that our models only use $2$ layers and require learning few
parameters.
Note that better results than those reported in Table~\ref{table:cifar} have
been obtained in the literature by using either data augmentation (around
$90\%$ on CIFAR-10 for~\cite{goodfellow2013,wan2013}), or external data (around
$70\%$ on STL-10 for~\cite{swersky2013}). We are planning to investigate similar data manipulations
in the future.

\begin{table}[hbtp]
   \centering
   \footnotesize
   \renewcommand\tabcolsep{0.19cm}
   \begin{tabular}{|c||c|c|c|c|c|c|c||c|c|c|}
      \hline
      Method & \cite{coates2011b} & \cite{sohn2012} & \cite{goodfellow2013} & \cite{coates2011} & \cite{bo2013} & \cite{gens2012} & \cite{zeiler2013} & CKN-GM & CKN-PM & CKN-CO\\
      \hline
      CIFAR-10  & 82.0 & 82.2 & \textbf{88.32} & 79.6 & NA   & 83.96 & 84.87& 74.84 & 78.30 & 82.18\\
      \hline
      STL-10  & 60.1 & 58.7 & NA    & 51.5 & \textbf{64.5} & 62.3 & NA& 60.04 & 60.25  & 62.32\\
      \hline
   \end{tabular}
   \vsb
   \caption{Classification accuracy in~$\%$ on CIFAR-10 and STL-10 without data augmentation.}\label{table:cifar}
\end{table}

\vsb

\vs
\section{Conclusion}\label{sec:ccl}
\vsb
In this paper, we have proposed a new methodology for combining kernels
and convolutional neural networks. We show that  
mixing the ideas of these two concepts is fruitful,
since we achieve near state-of-the-art performance 
on several datasets such as MNIST, CIFAR-10, and STL10, with simple architectures
and no data augmentation.
Some challenges regarding our work are left open for the future. The first one
is the use of supervision to better approximate the kernel for the prediction
task. The second consists in leveraging the kernel interpretation of our
convolutional neural networks to better understand the theoretical properties
of the feature spaces that these networks produce.

\vsb
\subsubsection*{Acknowledgments}
\vsb
This work was partially supported by grants from ANR (project MACARON ANR-14-CE23-0003-01),
MSR-Inria joint centre, European Research Council (project ALLEGRO), CNRS-Mastodons program (project GARGANTUA), and the LabEx PERSYVAL-Lab (ANR-11-LABX-0025).

\newpage
\small{
   \bibliographystyle{plain}
   \bibliography{abbrev,main}
}
 \appendix
 \section{Positive Definiteness of~$K$}\label{sec:appendixA}
To show that the kernel~$K$ defined in~(\ref{eq:kernel}) is positive definite
(p.d.), we simply use elementary rules from the kernel literature described in
Sections 2.3.2 and 3.4.1 of~\cite{shawe2004}.  A linear combination of p.d. kernels with non-negative weights is also p.d. (see Proposition 3.22
of\cite{shawe2004}), and thus it is sufficient to show that for all $\z,\z'$
in~$\Omega$, the following kernel on $\Omega \to \HH$ is p.d.:
\begin{displaymath}
   (\varphi,\varphi') \mapsto \big\|\varphi(\z)\big\|_\HH  \normH{\varphi'(\z')} e^{-\frac{1}{2\sigma^2} \normH{\tildephi(\z)-\tildephi'(\z')}^2}.
\end{displaymath}
Specifically, it is also sufficient to
show that the following kernel on $\HH$ is p.d.:
\begin{displaymath}
   (\phi,\phi') \mapsto \big\|{\phi}\big\|_\HH  \normH{\phi'} e^{-\frac{1}{2\sigma^2} \normH{\frac{\phi}{\|\phi\|_\HH}-\frac{\phi'}{\|\phi'\|_\HH}}^2}.
\end{displaymath}
with the convention $\phi/\|\phi\|_\HH=0$ if~$\phi=0$.
This is a pointwise product of two kernels and is p.d. when each of the two
kernels is p.d. The first one is obviously p.d.: $(\phi,\phi') \mapsto
\|{\phi}\|_\HH  \normH{\phi'}$. The second one is a composition of the Gaussian
kernel---which is p.d.---, with feature maps $\phi/\|\phi\|_\HH$ of a
normalized linear kernel in~$\HH$.  This composition is p.d. according to
Proposition 3.22, item (v) of~\cite{shawe2004} since the normalization does
not remove the positive-definiteness property.

\section{List of Architectures Reported in the Experiments}\label{appendix:arch}
We present in details the architectures used in the paper in Table~\ref{table:arch}.
\begin{table}[hbtp]
   \centering
   \begin{tabular}{|*{9}{c|}}
      \hline
      Arch. & $N$ & $m_1$  & $p_1$  &  $\gamma_1$ & $m_2$ &  $p_2$ & $S$  &  $\sharp$ param\\
      \hline
      \hline
      \multicolumn{9}{|c|}{MNIST} \\
      \hline
      CKN-GM1 & 2 &  $1 \times 1$  &  12  & 2 &  $3 \times 3$ &  50 &  $4 \times 4$ & $5\,400$\\
      \hline
      CKN-GM2 & 2 &  $1 \times 1$  &  12  & 2 &  $3 \times 3$ &  400 &  $3 \times 3$& $43\,200$ \\
      \hline
      CKN-PM1 & 1 &  $5 \times 5$  &  200  & 2 &  - &  - &  $4 \times 4$  & $5\,000$ \\
      \hline
      CKN-PM2 & 2 &  $5 \times 5$  &  50  & 2 &  $2 \times 2$ &  200 &  $6 \times 6$ & $41\,250$ \\
      \hline
      \hline
      \multicolumn{9}{|c|}{CIFAR-10} \\
      \hline
      CKN-GM & 2 &  $1 \times 1$  &  12  & 2 &  $2 \times 2$ & 800 &  $4 \times 4$ & $38\,400$\\
      \hline
      CKN-PM & 2 &  $2 \times 2$  &  100  & 2 &  $2 \times 2$ &  800 &  $4 \times 4$ & $321\,200$\\
      \hline
      \hline
      \multicolumn{9}{|c|}{STL-10} \\
      \hline
      CKN-GM & 2 &  $1 \times 1$  &  12  & 2 &  $3 \times 3$ & 800 &  $4 \times 4$ & $86\,400$\\
      \hline
      CKN-PM & 2 &  $3 \times 3$  &  50  & 2 &  $3 \times 3$ &  800 &  $3 \times 3$ & $361\,350$\\
      \hline

   \end{tabular}
   \caption{List of architectures reported in the paper. $N$ is the number of layers; $p_1$ and~$p_2$ represent the number of filters are each layer; $m_1$ and~$m_2$ represent the size of the patches~$\NN_1$ and~$\NN_2$ that are of size~$m_1 \times m_1$ and~$m_2 \times m_2$ on their respective feature maps~$\zeta_1$ and~$\zeta_2$; $\gamma_1$ is the subsampling factor between layer 1 and layer 2; $S$ is the size of the output feature map, and the last column indicates the number of parameters that the network has to learn.}
   \label{table:arch}
\end{table}

\end{document}